\documentclass[sigconf]{acmart}

\usepackage{caption}

\AtBeginDocument{%
  \providecommand\BibTeX{{%
    \normalfont B\kern-0.5em{\scshape i\kern-0.25em b}\kern-0.8em\TeX}}}

\setcopyright{acmcopyright}
\copyrightyear{2020}
\acmYear{2020}
\acmDOI{10.1145/1122445.1122456}

\acmConference[ICONS '20]{ICONS 20: International Conference on Neuromorphic Systems}{June 03--05, 2018}{Woodstock, NY}
\acmBooktitle{Woodstock '18: ACM Symposium on Neural Gaze Detection,
  June 03--05, 2018, Woodstock, NY}
\acmPrice{15.00}
\acmISBN{978-1-4503-XXXX-X/18/06}

\newcommand{\NW}[1]{}
\newcommand{\FF}[1]{}
\newcommand{\PB}[1]{}
\newcommand{\NeG}[1]{}
\newcommand{\ZZ}[1]{}

\begin{document}

\title{Towards On-Chip Bayesian Neuromorphic Learning}

\author{Nathan Wycoff}
\email{nathw95@vt.edu}
\affiliation{%
  \institution{Department of Statistics, \\Virginia Tech}
}

\author{Prasanna Balaprakash}
\email{pbalapra@anl.gov}
\affiliation{%
  \institution{Math and Computer Science Division, \\ Argonne National Laboratory}
}

\author{Fangfang Xia}
\email{fangfang@anl.gov}
\affiliation{%
  \institution{Data Science and Learning Division, \\ Argonne National Laboratory}
}

\renewcommand{\shortauthors}{Wycoff et al.}

\begin{abstract}

If edge devices are to be deployed to critical applications where their decisions could have serious financial, political, or public-health consequences, they will need a way to signal when they are not sure how to react to their environment \NeG{4 uses of they/their, I might remove first their and replace last with the}. For instance, a lost delivery drone could make its way back to a distribution center or contact the client if it is confused about how exactly to make its delivery, rather than taking the action which is ``most likely" correct. This issue is compounded for health care or military applications. However, the brain-realistic \NeG{neurobiologically realistic?} temporal credit assignment problem neuromorphic computing algorithms have to solve is difficult. The double role \NeG{which/that} weights play in backpropagation-based-learning, dictating how the network reacts to both input and feedback, needs to be decoupled. \texttt{e-prop 1} \NeG{I might cite, even in abstract} is a promising learning algorithm that tackles this with Broadcast Alignment (a technique where network weights are replaced with random weights during feedback) and accumulated local information. We investigate under what conditions the Bayesian loss term can be expressed in a similar fashion, proposing \NeG{expressED, investigaTE, proposing -> propose, following "We"}an algorithm that can be computed with only local information as well \NeG{remove as well} and which is thus \NeG{remove is thus} no more difficult to implement on hardware. This algorithm is exhibited on a store-recall problem, which suggests that it can learn good uncertainty \NeG{learn good uncertainty -> approximate uncertainty?} on decisions to be made over time.

\FF{Some points I would try to hit: Getting temporal credit assignment right in a brain realistic way is hard. The double role weights play in backprop needs to be decoupled. \texttt{e-prop 1} is a promising learning algorithm that tackles this with BA and accumulated local information. The important uncertainty aspect however is not addressed. It's unclear that the variational loss term can be expressed in a similar local fashion. We try to explore the possibility of implementing Bayesian learning on chip. We propose an algorithm that can be computed with local information as well and thus no more unrealistic. We show the results in a simple store-retrieve problem where uncertainty was expressed as expected. This suggests potential utility in ... and future direction for ...}
\end{abstract}

\ccsdesc[500]{Computing methodologies~Machine learning algorithms}
\ccsdesc[500]{Hardware~Neural systems}
\ccsdesc[500]{Mathematics of computing~Variational methods}

\keywords{spiking neural networks, variational inference, bayesian statistics, neuromorphic computing}

\begin{teaserfigure}
  \includegraphics[width=\textwidth]{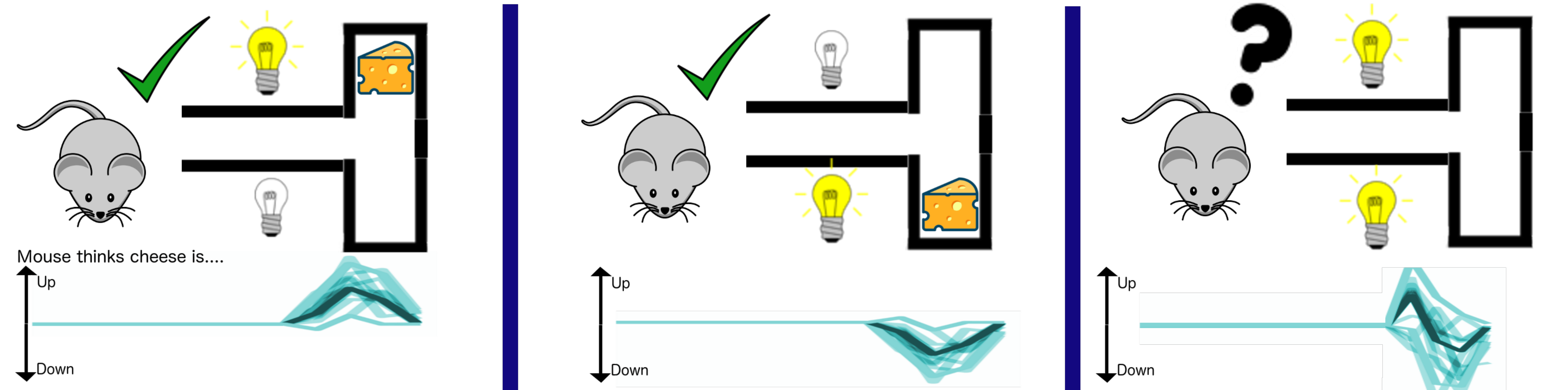}
  \caption{A mouse learns to associate a visual signal at the beginning of the experiment with the location of the reward at the end, a task an artificial Spiking Neural Network can solve as well. This article \NeG{Should be consistent between "We" or "This article" usage} introduces a method to allow the learning agent to express confusion when presented with a novel situation, such as both visual signals being triggered.}
  \label{fig:teaser}
\end{teaserfigure}

\maketitle

\section{Introduction}

AI algorithms have made ridiculous \NeG{promising, exciting, brilliant?} progress in recent years at solving well-defined problems with high accuracy, but are generally not capable of determining whether a particular problem is solvable with the data they have been given. Classical AI algorithms will heroically ``solve" problems which they are completely unequipped to handle without complaint. One popular solution for Uncertainty Quantification (UQ) is Bayesian inference, a procedure applicable to any statistical AI model. \PB{I would merge this and previous paragraph and make it shorter!}\NW{Done!}

\NW{Neuromorphics is a solution for portability and energy consumption.}
Another of AI's \PB{NmC?}\NW{I think I'm phrasing it as NmC is solving a problem for AI} major challenges is a two-for-one \NeG{I like "two-fold"}: portability and energy consumption, a pair of problems to which neuromorphic computing offers a solution. Progress has been made in the transfer of classical artificial neural network weights to a spiking neural network (SNN), so that a relationship learned offline can be deployed on the edge. However, in order for an edge device to be adaptive, it would \NeG{past/present phrasing again -> it needs to} have to learn while under operation. It is therefore important \NeG{->appropriate} to partition neuromorphic algorithms into two groups: those which can learn on-chip, and those which can only be deployed on-chip.

\NW{[Bring uncertainty and portability together.]}
AI applications for which portability is essential have even greater need for UQ than most, as by definition a human may not be available to monitor the process. Indeed, as we describe in Section \ref{sec:bayesbg}, previous research has developed neuromorphic Bayesian algorithms. However, to the best of our knowledge, an algorithm for Bayesian neuromorphic learning \textit{on-chip} has not been described in the academic literature. 

To address this gap, we formulate a pseudo-gradient based algorithm which performs Variational Bayesian Inference, building on \texttt{e-prop 1} \cite{Bellec2019}, a recently developed algorithm for on-chip learning. This endeavor presents several important challenges. First, we cannot introduce any additional need for information that is not local to each neuron. Crucially, as we show in Section \ref{sec:methods}, for a certain class of prior distributions, this is indeed the case. Second, Broadcast Alignment \NeG{italicize?}, the information locality method employed by \texttt{e-prop 1}, has \NeG{has->provides} little in the way of theoretical guarantees, and has been found to be unsuitable for certain applications \cite{Bartunov2018}. We find that it still enables learning in a store-recall experiment (Section \ref{sec:experimentResults}), but with less efficiency than when full weight transport is enabled. Fortunately, it still displays useful UQ behavior, showing high uncertainty when presented with a previously unseen situation.

\PB{Write a short summary on what exactly you did;}


%
%
%
%
%

\section{Background}


\subsection{Neuromorphic Learning on Chip}

This article builds on a framework for neuromorphic learning described in \cite{Bellec2019} called \textit{e-prop 1}, which purports to be deployable for on-chip learning. It employs Broadcast Alignment (BA) \cite{Samadi2017}, a variant of Feedback Alignment \cite{Lillicrap2016,Nokland2016}, to overcome the weight transport problem \cite{Grossberg1987} and develops a reinforcement-learning-like eligibility trace to avoid the future errors problem \cite{Lillicrap2019}. 
\FF{This is a succinct summary. However, it's challenging for readers to digest all these concepts at once. If space could be save from the introduction section, it might be beneficial to give some intuition of what all this means. Might also be useful to cite [Backpropagation and the Brain] to denote where e-prop sits in the synaptic change spectrum in Fig 1}

\NW{I haven't taken a close enough look at the article you mention to feel comfortable citing it (at least today, perhaps in the revision), though definitely plan to look at it more carefully soon. I've added the below para to try to give just a little bit extra intuition on what eprop is doing, but I'm not 100\% satisfied with it. I'm finding this section difficult to write.}

Intuitively, \texttt{e-prop 1} doesn't explicitly consider future contributions to the error, but does so implicitly by storing information about past interactions in its eligibility trace. As demonstrated in \cite{Bellec2019} and reproduced in Section \ref{sec:experiment}, this is sufficient to solve problems with delayed reward. 

A neuron's eligibility trace is characterized by its activity discounted over time. This information, encoding past behavior, is combined with information about present error to give a weight update. While \texttt{e-prop 1} gives similar results to past online learning algorithms for simple neuronal dynamics, its power \NeG{advantage} lies in its generalizability \NeG{portability} to SNNs built on arbitrary neuron models. 

In brief, \texttt{e-prop 1}  works by separating the derivative of the total loss $l$, into two components at each of $T$ time steps:

\begin{equation}
    \frac{d l}{d \theta} = \sum_{t=1}^T \frac{d l}{d z_t} \frac{d z_t}{d \theta}
\end{equation}{}

where here, $z_t$ represents the observable state of some neuron (for a spiking neuron, this is typically simply whether it was spiking at time $t$ or not), and $\theta$ represents a network weight. The second component in the sum is the eligibility trace, which tracks past behavior of the neuron, and may be calculated on-line, see \cite{Bellec2019} for details. The first component represents the contribution of that neuron to the error, and cannot be calculated on-line. The characteristic approximation of \texttt{e-prop 1}  is to replace $\frac{d l}{d z_t}$ with $\frac{\partial l}{\partial z_t}$; in words: replacing the total error contribution over the entire future trial period with the instantaneous contribution. Alternatives to this are explored in \cite{Bellec2019}.

\subsection{From Classical to Variational Inference}

In classical neural network inference, we find a set of weights, $\theta$, that minimizes loss, perhaps subject to regularization. In the standard variational approach, we no longer are optimizing individual weights; instead, each weight $\theta_{i,j}$ has a mean parameter $\lambda_{i,j}$ and variance parameter $\phi_{i,j}^2$. These \textit{variational parameters} are optimized, just as the weights themselves would be, to minimize a trade-off between data misfit and regularization.
Specifically, whereas a particular set of weights maps to a well defined loss on a particular dataset, a particular set of variational parameters leads to a distribution on losses. We thus minimize the expected loss. 
Regularization of the variational distribution, denoted $q_{\lambda, \phi}(\theta)$, is achieved by penalizing its KL distance from a given prior, denoted by $p(\theta|\psi)$. Our variational cost is therefore:

\begin{equation}
    l(\lambda,\phi) = -\underset{\theta \sim q_{\lambda,\phi}}{\mathbb{E}} [\log L(\mathbf{y}|\theta, \mathbf{X})] + KL(q_{\lambda,\phi}(\theta)||p(\theta|\psi))
    \label{eq:popkl}
\end{equation}{}

\FF{Consistency in using log likelihood?}\NW{Thanks!} Though the KL divergence between two multivariate Gaussian distributions is available in closed form, the expected loss term in Equation \ref{eq:popkl} is not. However, notice that the expectation is with respect to the variational distribution, which is easy to sample from. Thus, we can form a Monte-Carlo estimate of our variational loss:

\begin{align}
     & l(\lambda, \phi) \approx
     \frac{1}{V} \sum_{v=1}^V  [-\log L(y|\theta_v, \mathbf{X})] + KL(q_{\lambda, \phi}(\theta)||p(\theta|\psi)) \\ 
     & = \frac{1}{V} \sum_{v=1}^V  [-\log L(y|\theta_v, \mathbf{X})] + 
     \frac{1}{2}[\sum_{i,j} \frac{\phi_{i,j}^2}{\psi} +
     \sum_{i,j} \frac{\psi}{\phi_{i,j}^2} + \sum_{i,j} \frac{\lambda_{i,j}^2}{\phi_{i,j}^2}]
\end{align}{}

Here, $\psi$ is the prior variance hyperparameter, which plays the roll of the $\ell_2$ regularization coefficient in classical machine learning. \FF{$B$ or $V$? Is $l$ overloaded?}\NW{Thanks!}

\section{Methods}

\label{sec:methods}

\texttt{e-prop 1} provides us with a way of estimating the gradient of each individual $\log L(y|\theta_v, \mathbf{X})$ term, which can be averaged to give an estimate of the overall loss gradient \footnote{However, as shown in \cite{Blundell2015}, this estimate is not unbiased. It is likely that we could do better by incorporating the changes suggested in that article. This task is left as future work.}. This mean loss gradient is combined with the prior-KL gradient. For the variational mean, this is given by:

\begin{equation}
    \frac{\partial KL(q_{\lambda, \phi}(\theta)||P(\theta))}{\partial \lambda_{i,j}} = 
    \frac{\lambda_{i,j}}{\phi_{i,j}^2}
\end{equation}{}

and that for the variational signed standard deviation by\footnote{Most authors parameterize the variational variance in log space, so that no positivity constraints are required. We elect to use a standard deviation parameterization instead, where possible negatives disappear after squaring.}:

\begin{equation}
    \frac{\partial KL(q_{\lambda, \phi}(\theta)||P(\theta))}{\partial \phi_{i,j}} = 
    \phi_{i,j} (\frac{1}{\psi} - \frac{1}{\phi_{i,j}^2})
\end{equation}{}

Notably, the prior gradients for the variational parameters on the weight between $i,j$ do not depend on those for any other $l,k$ pair.\FF{Have we made clear $i,j$ are neurons? Perhaps $u,v$ instead of $l,k$?} \PB{Not clear to me; please explain} \NW{Whoops, there was a notation error that I have fixed. Are there clarity issues beyond that?}Thus, the gradient for the entire posterior-KL term may be expressed as a sum of gradients involving only local information, and thus is local itself. This demonstrates that the algorithm presented here does not aggravate the locality issue. \PB{so you still have to do gradient descent? How long? how the catostrophic forgetting will be handled if you keep updating the parameters in an online setting?} \NW{Absolutely, still iteratively optimizing loss. I'm not sure what you mean by "how long?", are you saying I should say how many epochs I did training for here? This method does not address catastrophic forgetting.}
\FF{The fact that the proposed algorithm is no less local is an important point that deserves to be in the abstract. Also consider foreshadowing that in the introduction with something like: we explore whether KL could also be expressed using only local information...} 

\section{Experiment}

\subsection{Experimental Setup}

\label{sec:experiment}

In this section, we will \NeG{"we will" tense is not consistent throught section, sorry to be stickler on this! very minor tho} evaluate our algorithm on a simplified version of the temporal credit assignment task considered using SNNs in \cite{Bellec2019} and  model mice \PB{model?}\NW{good call}in \cite{Morcos2016,Engelhard2019}. Our virtual mouse ``perceives light" through 20 input spike channels, ten for up and ten for down. When a light is turned on, each of the input neurons associated with it fire. Then, there is a break period, with no input, and, finally, a ``cue" signal, during which an additional 10 channels fire, telling the virtual mouse it needs to make a decision: is the cheese Up or Down? That the SNN has to make a decision based on past events makes this problem interesting, and even impossible for certain neural network architectures, such as non-recurrent classical neural networks or simple Leaky Integrate and Fire (LIF) SNNs \PB{revisit this; did not read well!}\NW{Thanks, I have given it another look.}. \FF{Looks fine to me now.}

As such, following \cite{Bellec2019}, we will use LIF neurons with an adaptive threshold: the potential required to generate a spike increases immediately after an action potential, and then decays exponentially back to the baseline
\footnote{\PB{Why this is in footnote; should be in the main text; you cannot skip the setup; Where was this implemented and how?} \FF{I understand this was probably done for space concern? I would try to squeeze this back to the main text, at the cost of cutting introduction text and even some text introducing variational inference or related work.} \NW{Here's a hybrid, where most info has been moved to the body, aside from the specifics of the neuronal dynamics}(broadly following \cite{Bellec2019}): The membrane potential decay constant is 20 ms and the threshold decay constant is 2000 ms for all neurons. Each neuron triggers an action potential when its membrane potential crosses 1, and each spike increases the threshold by 2 (before exponentially decaying back to 1).}. 
We used gradient descent with a learning rate of 5e-4, a mini-batch size of 20, 5 variational samples ($V = 5$), and 5000 epochs. Our SNN had $H=100$ neurons featuring all-to-all connectivity but no connections allowed between a neuron and itself. The simulation is run using time steps of 50 ms, which is also the duration of the input signal. Consequently, all input channels fire at once, on the second time step. The period between the input signal and the cue is sampled randomly and uniformly between 500 ms and 1,500 ms, at which point the cue period begins, which lasts 150 ms. All 10 cue neurons fire at random during one and only one of these time steps. We set the prior variance $\psi$ to $\frac{1}{H} = 0.01$.

\subsection{Results}

\label{sec:experimentResults}

\begin{figure}[ht]
    \centering
    \includegraphics[scale=0.67]{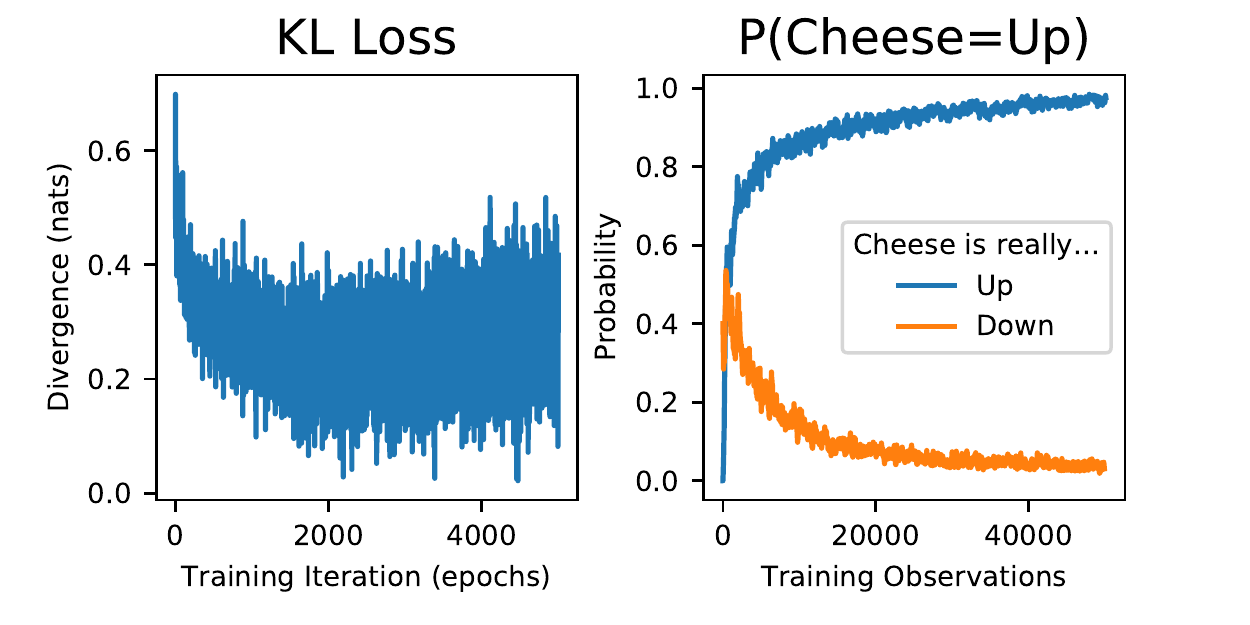}
    \includegraphics[scale=0.67]{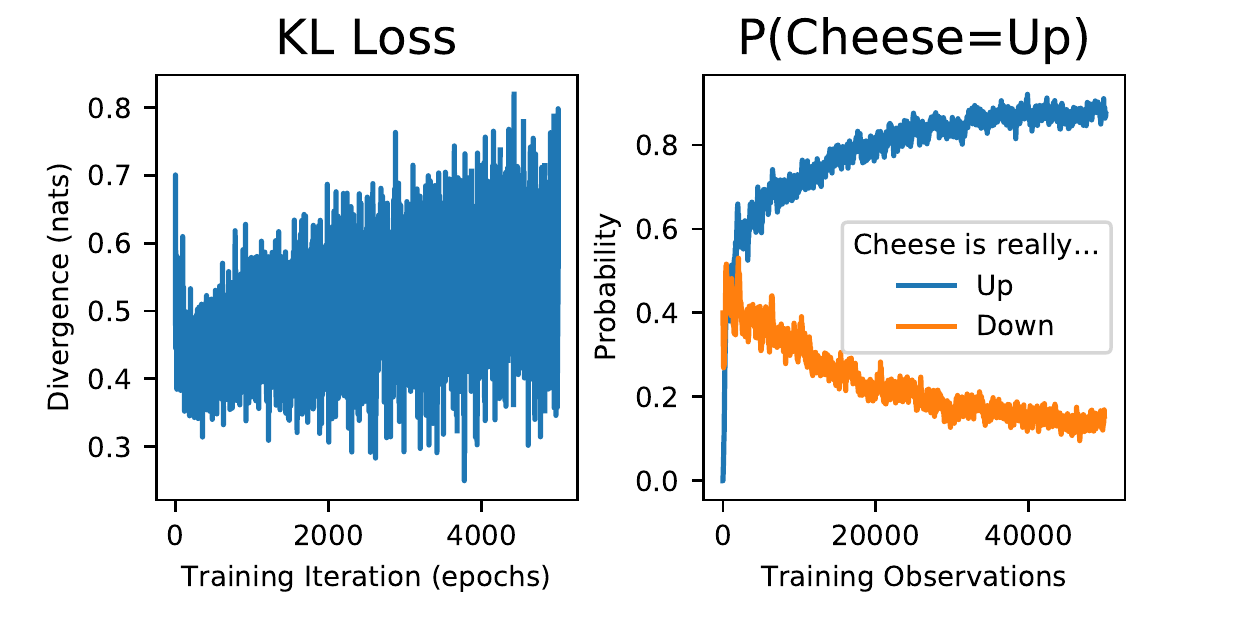}
    \caption{\textmd{\textbf{Top Row:} training with weight transport. \textbf{Bottom Row:} avoiding weight transport via Broadcast Alignment. \textbf{Left Column:} Following the pseudo-gradients described above iteratively reduces KL divergence despite high variance when weight transport is allowed. With BA, we see an initial drop in KL divergence, followed by oscillations in the cost. \textbf{Right Column:} The SNN learns the task in both cases, though higher accuracy is achieved without BA. The two lines give the SNN's probability that the cheese is up when it really is (blue) or when it really is down (orange), smoothed via an exponential moving average. An ideal learner quickly learns to separate Up and Down instances. \FF{I tried to lose some of the bold font; would this be better for long captions?}\NW{Good idea, I put it on the other figure too. I don't understand why there's so much space under this image...}}}
    \label{fig:training}
\end{figure}

As shown in Figure \ref{fig:training}, the Bayesian SNN learns to classify Up versus Down \NeG{Italicize Up/Down} input channels over the course of the training whether weight transport is allowed or not. However, the KL cost does not decrease as reliably when BA is used, and the training set performance, as measured by the probability of giving the correct direction, is also worse when BA is employed.  \PB{can you explain the results.} 

\begin{figure}[ht]
    \centering
    \includegraphics[scale=0.67]{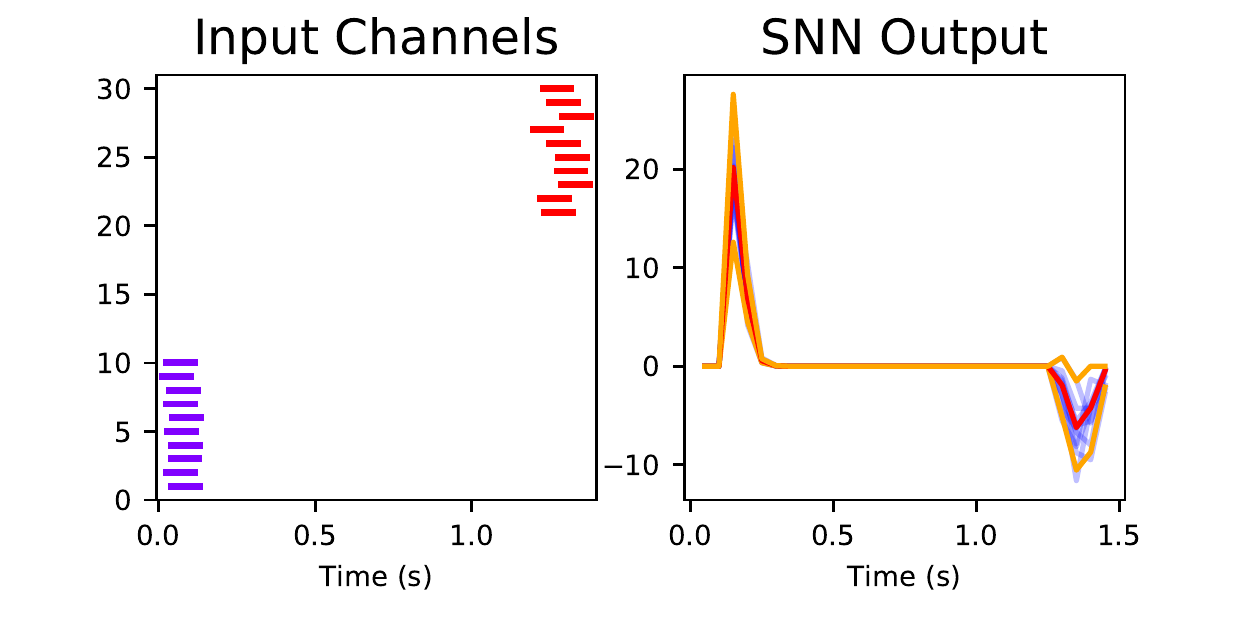}
    \includegraphics[scale=0.67]{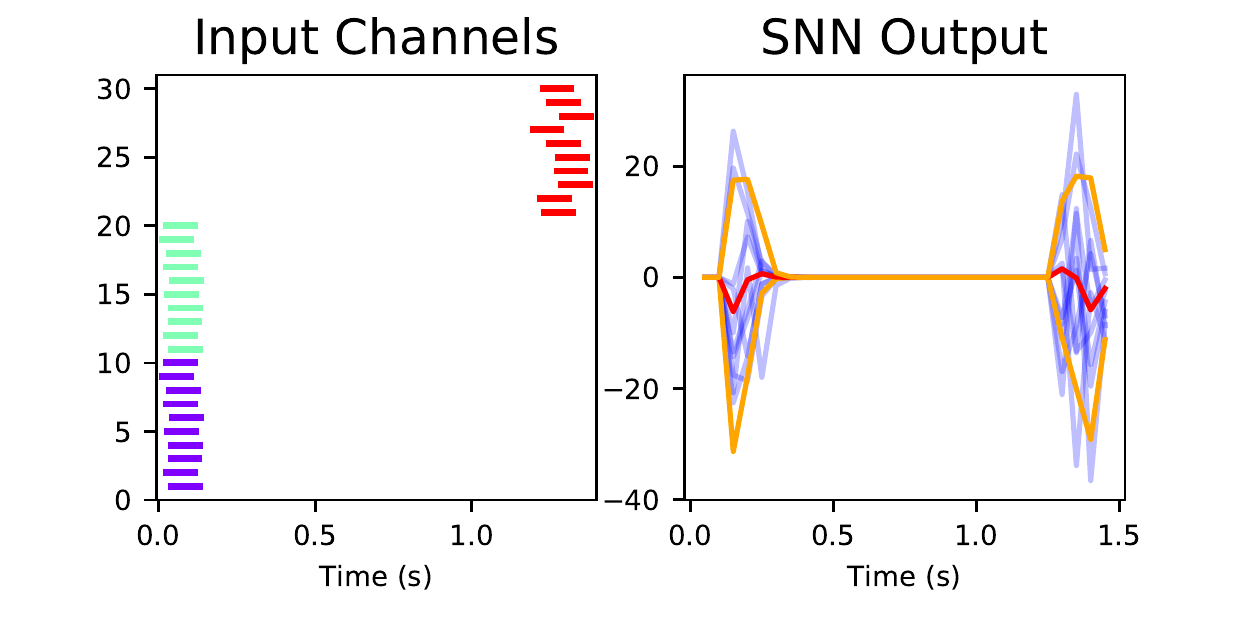}
    \caption{\textmd{\textbf{Top Row:} the Bayesian SNN has correctly identified that channels 1-10 firing mean that the value stored needs to be negative (not shown: 11-20 cause it to give positive values). \textbf{Bottom row:} the Bayesian SNN is presented with both channels on, a novel situation, and shows high uncertainty in its response, as desired. The \textbf{left column} gives input action potentials. The \textbf{right column} illustrates the posterior distribution through a Monte Carlo sample therefrom, with 10\textsuperscript{th} and 90\textsuperscript{th} quantiles shown in orange, the median shown in red, and 10 sample trajectories in faded blue.}}
    \label{fig:output}
\end{figure}

\FF{Consider moving this up above the previous paragraph. Limitations could be in the last discussion paragraph} Figure \ref{fig:output} demonstrates that the Bayesian SNN learns to remember which of Up or Down was signalled, a task the vanilla \texttt{e-prop 1}  algorithm can tackle as well. The Bayesian SNN, however, also gracefully handles the case where both input signals are simultaneously presented (a situation not present in any of the training cases) by giving high uncertainty as to whether the cheese is Up or Down. 
\FF{If this subsection is short, we might consider changing EXPERIMENT to RESULTS.}\NW{I have added some stuff and moved some stuff around}. 

{}

{}

\subsection{Limitations}

We have used minibatching in order to reduce gradient variance. \PB{so these are limitations of the study? separate that from the results..} \NW{Done!} This is not compatible with online learning, so smaller learning rates or variance reduction techniques will be required. Similar issues accompany our sampling and evaluating $V = 5$ weights from the varitional distribution for each element of the minibatch.
\NeG{While I agree with PB to separate, perhaps include in conclusion rather than having separate section?}


\section{Related Work}

Variational inference in the context of classical neural networks has received significant attention \cite{graves2011,hinton1993,Blundell2015}. 
\label{sec:bayesbg}
It has also been studied in the context of spiking neural networks. \cite{Rezende2011} examined how Variational Inference could be used to infer states of unobserved neurons given some set of observed neurons for which spike trains are known. They also showed that the STDP learning rule could be derived from that model, though left discussion of biological plausibility (and, hence, we would argue, neuromorphic plausibility) as future work. In order for neuromorphic hardware to carry out the algorithm we outline, it will need to have access to standard Gaussian random numbers. \cite{Yang2020} demonstrate how to generate these variates using Spintronics by first summing uniform random variables, which are also provided by Loihi. 

\section{Conclusions and Research Directions}

\label{sec:conclusion}

This article has demonstrated a first step on the road to success in on-chip Bayesian neuromorphics. We learned the posterior of a small, recurrently connected SNN trained on a simple store-recall problem using variational inference. The network was able to accurately solve the task, as well as demonstrate uncertainty when asked to solve a problem it was not equipped for. 

The ultimate goal of this research is to run this algorithm on a neuromorphic chip with learning enabled, such as Loihi. To make this possible, we will need to demonstrate learning with both a minibatch and variational sample size of 1. 

Further, while the variational approximation that we used seemed \NeG{seems} to be suitable for this task, it has been shown that some classical architectures exhibit pathological behavior when using the mean field approximation \cite{Foong2019}. Exploration of when the approximation is appropriate in the context of SNNs we leave as future work.

Some neuromorphic chips, such as Loihi, have the ability to simulate more complicated neuron models, such as those with multiple compartments. Potentially, this hardware could be enlisted to instead \NeG{remove instead?} perform side-by-side simulation of the same network with different samples of its weights, a ``multi-channel minibatching", lifting our requirement that $V = 1$. 

%
%
%
%

\begin{acks}
Part of this work is supported by the Laboratory Directed Research \& Development Program at Argonne National Laboratory. 
The authors thank Neil Getty and Zixuan Zhao for helpful comments and discussion. 
\end{acks}

\bibliographystyle{ACM-Reference-Format}
\bibliography{beprop}

\appendix

\end{document}